\pdfoutput=1
\documentclass[10pt,twocolumn,letterpaper]{article}
\usepackage[pagenumbers]{iccv}
\usepackage{times}
\usepackage{latexsym}
\usepackage[T1]{fontenc}
\usepackage[utf8]{inputenc}
\usepackage{microtype} % compact stuff
\usepackage{inconsolata} % aesthetiic
\usepackage{amsmath}
\usepackage{amsfonts}
\usepackage{pgfplots}
\usepackage{graphicx}
\usepackage{amssymb}
\usepackage{alphabeta}
\usepackage{tabularx}
\usepackage{multicol}
\usepackage{makecell}
\usepackage{fancyvrb}
\definecolor{iccvblue}{rgb}{0.21,0.49,0.74}
\usepackage[pagebackref,breaklinks,colorlinks,allcolors=iccvblue]{hyperref}

\DeclareUnicodeCharacter{2212}{−}
\usepgfplotslibrary{groupplots,dateplot}
\usetikzlibrary{patterns,shapes.arrows}
\pgfplotsset{compat=newest}

\DeclareMathOperator*{\argmax}{arg\,max}

\DeclareMathOperator*{\softmax}{softmax}

\title{Understanding the Quality-Diversity Trade-off in Diffusion Language Models}

\author{Zak Buzzard \\ University of Cambridge \\ \texttt{zzb20@cam.ac.uk}}

\begin{document}
\maketitle

\begin{abstract}
% - NAR, diffusion text on the up
% - Study the diversity/quality trade-off
Diffusion models have seen immense success in modelling continuous data across a range of domains such as vision and audio. Despite the challenges of adapting diffusion models to discrete data, recent work explores their application to text generation by working in the continuous embedding space. However, these models lack a natural means to control the inherent trade-off between quality and diversity as afforded by the temperature hyperparameter in autoregressive models, hindering understanding of model performance and restricting generation quality. This work proposes the use of classifier-free guidance and stochastic clamping for manipulating the quality-diversity trade-off on sequence-to-sequence tasks, demonstrating that these techniques may be used to improve the performance of a diffusion language model.
\end{abstract}
%   and uses these methods to produce state-of-the-art generation quality using a diffusion language model.

% General
\newcommand{\N}{\mathcal{N}}
\newcommand{\nth}[1]{{#1}_\text{th}}

% Specific
\newcommand{\x}{{\bf y}}
\newcommand{\cond}{{\bf x}}
\newcommand{\Lo}{\mathcal{L}}
\newcommand{\Ls}{\mathcal{L}_\text{simple}}
\newcommand{\Li}{\mathcal{L}_\text{IS}}
\newcommand{\La}{\mathcal{L}_\text{anchor}}
\newcommand{\Emb}[1]{\text{Emb}\left(#1 \right)}
\newcommand{\w}{{\bf w}}
\newcommand{\condw}{{\bf v}}
\newcommand{\wspace}{\mathcal{W}}

\newcommand{\cfg}{\text{cfg}}
\newcommand{\clamp}{\text{clamp}}
\newcommand{\predc}{\hat\x_0^c}
\newcommand{\predu}{\hat\x_0^u}
\newcommand{\pred}{\hat\x_0}
\newcommand{\predtilde}{\widetilde{\x}_0}

\newcommand{\name}{GuideDiffuSeq}

\section{Introduction}
% How to justify this project?
% - application to seq2seq tasks

% Diffusion models \cite{DDPM} have emerged in recent years as an important class of generative models, with diverse applications across a range of domains including image and audio. Diffusion models are trained to reverse a forward noising process, enabling their use as a generative model when applied iteratively to pure noise.
Diffusion models \cite{DDPM} have emerged in recent years as an important class of generative model, achieving state-of-the-art results across a range of generative tasks. Despite significant successes on the tasks of image \cite{diffusion_beats_gans, latent_diffusion_models, DDIM, GLIDE, classifier_free_guidance} and audio \cite{DiffWave_audio_diffusion, WaveGrad_audio_diffusion} synthesis, diffusion models face challenges when applied to natural language due to its discrete nature. Current approaches to this task fall into two categories: discrete variants of the diffusion model \cite{D3PM, reparameterized_discrete_diffusion_model}, applied directly to the tokens of a text sequence, or continuous parameterisations applied to the continuous embedding space \cite{diffusionlm, diffuseq, difformer, SeqDiffuSeq}. Working in the embedding space enables the use of semantic data encoded in token embeddings, yet the sparsity of this space poses challenges: the diffusion model is restricted in its generation to produce items in the vicinity of a token's embedding.

Despite these drawbacks, several works have achieved near state-of-the-art performance on tasks such as text summarisation and neural machine translation \cite{difformer, SeqDiffuSeq}, highlighting this area's potential. Furthermore, as a non-autoregressive text generation method, diffusion models are free from the inherent drawbacks of autoregressive generation such as search errors and bias towards shorter generations (when using beam search) \cite{cat-got-your-tongue, map-all-you-need}, although they are still subject to some form of exposure bias \cite{diffusion_exposure_bias, diffusion_exposure_bias_2}.

Keen to show the viability of this field, generation quality has been prioritised in prior works. In such cases, it is easy to disregard generation \textit{diversity}. \citealp{gans_fall_short} warn of the dangers of quality-only analysis; one method may not be asserted as superior to another unless it outperforms on both quality and diversity, as it is generally possible to trade one of these metrics off for the other. Accordingly, it is necessary to evaluate model quality across a range of diversities to gain understanding of the model's true `performance', and to make justified assertions regarding model superiority. It is therefore essential for language models to possess effective methods for manipulating this trade-off, enabling versatile usage at any chosen level of diversity, as well as potentially unlocking greater generation quality at the cost of diversity.

While some works make efforts to evaluate model diversity as well as quality \cite{diffuseq}, or explore the effect of MBR \cite{SeqDiffuSeq}, diffusion language models currently lack effective means of manipulating this trade-off. This project's key contributions are as follows:
\begin{itemize}
    \item We propose classifier-free guidance and stochastic clamping as methods for controlling the quality-diversity trade-off in diffusion language models, and demonstrate their efficacy.
    \item By applying these techniques to a diffusion language model, we achieve highly competitive results on the paraphrasing task using the Quora Question Pairs (QQP) dataset despite training for just three hours, far less than similar works.
    \item We open-source our implementation\footnote{\texttt{https://github.com/zzbuzzard/GuideDiffuSeq}} which was written from scratch using the \verb|diffusers| library \cite{diffusers}, rather than adapting an existing image diffusion codebase as in many prior works \cite{diffusionlm, diffuseq, SeqDiffuSeq, difformer}, leading to a much smaller repository overall. 
\end{itemize}

% in the local .. [[] neighrobuhood]
% with diverse applications across a range of domains including image and audio. Diffusion models are trained to reverse a forward noising process, enabling their use as a generative model when applied iteratively to pure noise.

% Recent work has begun to explore the application of diffusion models to text generation

\begin{figure*}
    \centering
    \includegraphics[width=0.9\linewidth]{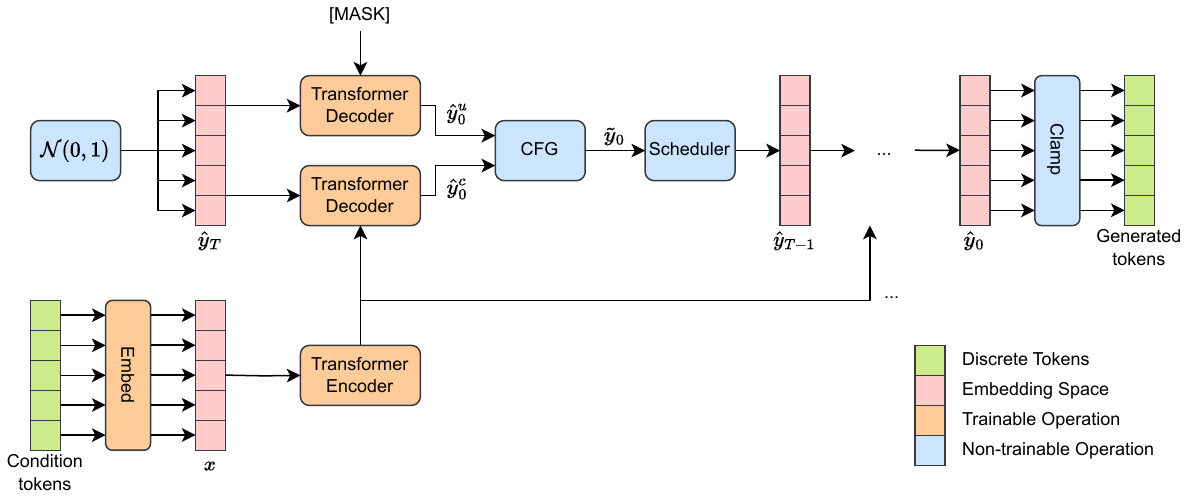}
    \caption{Summary of the proposed method: an embedding-space sequence-to-sequence diffusion model with a transformer encoder-decoder backbone, augmented with classifier-free guidance. The diffusion process generates a sequence of embeddings $\hat\x_0$ which are clamped to the nearest tokens.}
    \label{fig:big-diagram}
\end{figure*}

\section{Background}
\subsection{Diffusion Models}
To model a continuous distribution $q(\x)$, diffusion models define a forward process $q(\x_t \mid \x_{t-1})$ which gradually corrupts a sample $\x_0 \sim q(\x)$ over $T$ steps $\x_1, \x_2, ... \x_T$ (where $T$ is a hyperparameter) such that the distribution of samples at the $\nth{T}$ step approximates a tractable distribution, usually $\x_T \sim \N(0, 1)$. This is formulated as follows:
\[
    q(\x_t \mid \x_{t-1}) = \N(\x_t; \sqrt{1 - \beta_t} \x_{t-1}, \beta_t {\bf I}) \,,
\]
where hyperparameters $\beta_t \in (0, 1)$ indicate the variance scale for timesteps $1 \leq t \leq T$. This parameterisation has a convenient closed form 
\begin{align} \label{eq:closed_form}
q(\x_t \mid \x_0) = \N(\x_t; \sqrt{\overline{\alpha}_t} \x_0, (1-\overline{\alpha}_t) {\bf I}) \,,
\end{align}
using the standard notation $\alpha_t = 1-\beta_t$ and $\overline{\alpha}_t = \prod_{i=1}^t \alpha_i$. 

The distribution $q(\x_{t-1} \mid \x_t)$ is then parameterised by $p_\theta$ for steps $1 \leq t \leq T$, enabling the generative reverse process. In particular, $q(\x_{t-1} \mid \x_t)$ is modelled as a Gaussian distribution with learned mean and covariance:
\[
    p_\theta(\x_{t-1} \mid \x_t) = \N(\x_{t-1}; \mu_\theta(\x_t, t), \Sigma_\theta(\x_t, t))
\]
though it is common to fix the covariance to a series of time-dependent constants such as $\Sigma(\x_t, t)=\beta_t {\bf I}$ \cite{DDPM}. Following several text generation diffusion works \cite{diffusionlm, diffuseq, difformer}, we use an equivalent parameterisation in which a neural network $f_\theta$ is trained to directly output the initial sample $\x_0$, as the prediction for $\x_{t-1}$ may be obtained directly from $f_\theta(\x_t, t) = \hat{\x}_0$ via \cref{eq:closed_form}. 

\citealp{DDPM} derive a variational lower bound, but empirically demonstrate a simplified objective $\Ls$ to be equally performant:
\[
    \Ls = \mathbb{E}_{\x_0 \sim p(\x), t}[||f_\theta(\x_t, t) - \x_0||^2]
\]

Denoising diffusion probabilistic models \cite{DDPM} carry out inference by producing noisy samples $\x_T \sim \N(0, 1)$, then iteratively sampling the previous timestep $\hat{\x}_{t-1}$ using the predicted distribution $p_\theta$, though far more efficient sampling methods exist \cite{dpm_solver_plusplus, PNDM}.

\subsection{Sequence-to-Sequence Text Generation with Diffusion Models}
%  Recent methods tackle this in one of two ways: several works apply discrete diffusion models directly to textual tokens, using a forward process which discretely corrupts tokens (for example by masking) \cite{D3PM, reparameterized_discrete_diffusion_model}, while other approaches work in the 
Text data is inherently discrete, presenting challenges for the application of diffusion models. While discrete variants exist, this work focuses on modelling the continuous embedding space, in which $\x_0 = [\Emb{\w_1},~\Emb{\w_2}, ..., \Emb{\w_n}]$ for a sequence of $n$ tokens $\w_1, ... \w_n$ from some token space $\wspace$. Following generation of a continuous sample $\hat{\x}_0$, a clamping step is required to obtain discrete tokens $\hat{\w}_1, ... \hat{\w}_n$. The distribution $q$ is additionally conditioned on a textual input $\cond = [\Emb{\condw_1},...,\Emb{\condw_m}]$.

It is common to use a transformer \cite{transformer} to parameterise the function $f_\theta$, with some works using encoder-decoder \cite{difformer, SeqDiffuSeq} (with $\cond$ fed to the encoder without noise while the decoder receives the noised $\x_t$) and others an encoder-only architecture \cite{diffusionlm, diffuseq, diffusum} (where the concatenated sequence $\cond || \x_t$ is input).

\subsection{Classifier-free Guidance}
Classifier-free guidance (CFG) \cite{classifier_free_guidance}, is a method for guiding the diffusion process without requiring a separate classifier. For a conditional diffusion model $\hat{\x}_0 = f_\theta(\x_t, t, \cond)$, classifier-free guidance perturbs the predicted value as follows\footnote{This formulation is equivalent to that in the original paper, which uses an $\epsilon$-predicting parameterisation}:
\[
    \widetilde{\x}_0 = f_\theta(\x_t, t, \emptyset) + s \cdot (f_\theta(\x_t, t, \cond) - f_\theta(\x_t, t, \emptyset))
\]
for some guidance scale $s \geq 0$. Larger values of $s$ shift the prediction $\widetilde{\x}_0$ further from the unconditional prediction $f_\theta(\x_t, t, \emptyset)$ in the direction of the conditional prediction, causing the condition to have increased effect.

CFG has seen wide usage in the domain of images, and is understood to trade off diversity for fidelity \cite{GLIDE, latent_diffusion_models}. While classifiers have been used to guide the diffusion process in text generation \cite{diffusionlm}, to the best of my knowledge classifier-free guidance has not yet been explored.

% In the domain of text, classifiers have been used to guide the diffusion process as a method for controllable generation, encouraging the output to meet a constraint such as a series of part-of-speech tags \cite{diffusionlm}. However, to the best of my knowledge, prior work has not yet explored applying classifier-free guidance to the domain of text.
% A strength value of 0 corresponds to unconditional generation, and 1 to normal conditional generation.
% From proposal:
% Classifier guidance has already seen usage in the application of diffusion to NLP; for instance, Diffusion-LM uses classifier-guidance for generating text conditional on some structure (e.g. a parse tree, PoS tags) or other property (e.g. sentiment) \cite{diffusionlm}. For a conditional diffusion model $\mu(x_t, t) \approx x_{t-1}$, classifier-free guidance on a class $c$ replaces the mean prediction $\mu(x_t, t \mid c)$ with $(1+w)\mu(x_t, t \mid c) - w \mu(x_t, t)$, where $w$ is the guidance strength.
% , for a conditional model $p(x)$, provides the output $(1+w)p(x\mid c) - w p(w)$

% TODO: put a fat diagram in...

\section{Method}
This project trains a baseline diffusion language model in the embedding space, then investigates its quality-diversity trade-off. This section details the design choices and their justifications. We name my method `Guided Diffusion Sequence-to-sequence' (\name) due to its use of classifier-free guidance.
% \Cref{fig:big-diagram} illustrates the overall inference process.

% TO MENTION
%  - beta schedule: sqrt and linear, linear better
%  - torch_cluster, diffusers, transformers

\subsection{Architecture}
Following \citealp{difformer, SeqDiffuSeq}, we use a transformer encoder-decoder architecture. As the condition $\cond$ does not change during inference, the encoder output may be shared across all timesteps improving efficiency (\cref{fig:big-diagram}) - this is not possible using an encoder-only architecture. While some previous works use learned embeddings for the timestep $t$ and position, we observe no performance difference when using the standard sinusoidal embedding \cite{transformer} for both. Full architecture and optimisation details are given in \Cref{sec:hyperparameters}.

\subsection{Training}
\citealp{difformer} introduce an additional training objective named the anchor loss to jointly optimise the model parameters and discourage inter-embedding similarity:
%observe high similarity between learned embeddings, and introduce an auxiliary loss term, the anchor loss, to counter this:
\[
    \La = \mathbb{E}_{\x_0, t, \cond} \left[ -\log p(\w_0, ... \w_n \mid \hat\x_0) \right] \,,
\]
where $\hat{\x}_0 = f_\theta(\x_t, t, \cond)$ and the logits of $p$ are given by the distances of $\hat\x_0$ to the embeddings:
\[
    p(\w_i \mid \hat\x_{0, i}) = \softmax_{\w \in \wspace} || \Emb{\w_i} - \hat\x_{0, i} ||_2
\]
This NLL loss term encourages the embeddings to spread out, which is beneficial for the generation process.
% This NLL loss term encourages the embe
% log(p_theta(w | x0)) = 
% log(p(w1 | x0) p(w2 | x0) ...) = log(p(w1 | x0)) ...
% denotes the clamping probability distribution,
% \[
%     p_\theta(\w \mid \x) \propto D(\Emb{\w} - \x)
% \]
% for some distance metric $D$ ( use the L2-norm).

%\citealp{diffusionlm} observe that the magnitude of the loss function changes significantly with the timestep $t$, being 
Empirically, the magnitude of the loss function varies significantly with $t$, as larger values of $t$ correspond to noisier samples $\x_t$ and thus represent a more difficult task. We therefore follow \citealp{diffusionlm} in the use of importance sampling, in which the timestep $t$ is chosen non-uniformly via $t \sim p_\text{IS}$ to prioritise more challenging timesteps:
% ||f_\theta(\x_t, t) - \x_0||^2
\[
    p_\text{IS}(t) \propto \sqrt{\mathbb{E}_{\x_0 \sim p}\left[\left(\Ls + \La\right)^2\right]}
\]
and the loss values reweighted to give an overall loss function of
\[
    \Lo = \mathbb{E}_{t \sim p_\text{IS}, \x_0 \sim p}\left[\frac{\Ls + \La}{p_\text{IS}(t)}\right]
\]
Importance sampling removes the need for modified variance schedules $\beta_{1:T}$, such as the square-root schedule from \citealp{diffusionlm}, and we obtain best performance for a simple linear schedule.

To enable classifier-free guidance at inference time, each input text is randomly replaced with a single \verb|[MASK]| token with probability $p$ during training to represent the null condition $\emptyset$, forcing the model to make an unconditional prediction. Following discussion in \citealp{classifier_free_guidance}, $p=0.1$ is used.

Finally, \citealp{diffusionlm} observe that embedding magnitude correlates negatively with word frequency and \citealp{difformer} demonstrate that this is detrimental to performance. To address this issue, we normalise each embedding to have mean 0 and standard deviation 1 following each optimizer step.

\subsection{Inference}
Diffusion model inference is expensive compared to similar generative models such as generative adversarial networks or variational autoencoders, requiring multiple network evaluations. While a naive approach would require $T=1000$ evaluations per sample, a large body of work focuses on improving the efficiency of inference \cite{DDIM, PNDM, consistency}. We select the recent DPM++ scheduler \cite{dpm_solver_plusplus} for inference, enabling high quality samples to be produced in under 10 steps rather than requiring 100s or 1000s. All experiments in this project use either 20 or 5 inference steps.

While the length of the output $\x$ is known at train-time, it is not known at inference time when given a condition $\cond$. \citealp{diffuseq, SeqDiffuSeq} explicitly include the \verb|[EOS]| token, as well as padding tokens \verb|[PAD]|, in the training process, allowing the model to specify the length of its output by generating an \verb|[EOS]| token (followed by padding tokens). While this allows the model to choose the length itself, it comes with several drawbacks: model capacity is wasted on learning to output large numbers of \verb|[PAD]| tokens, and including padding in training and inference adds significant computational overhead (max output length in both papers is set to 128, whereas the average length is far less). Instead, we remove end-of-sentence and padding tokens from training, and choose the length explicitly, masking out padding from the transformer. This allows for length-controllable generation. \citealp{difformer} predict the length from the encoder output, allowing complex length predictions to be made at inference time.
% Additionally, during inference, the input sequence remains constant while the output sequence is progressively de-noised. It is therefore sufficient to compute the encoder output just once rather than recomputing at each timestep.

% \subsection{Improving Generation Quality}
%  - CFG(-lerp)
%  - clamping trick (/lerp)
%  - normalisation
%  - interaction between the two
% Diffusion models, like autoregressive models, are subject to exposure bias \cite{diffusion_exposure_bias, diffusion_exposure_bias_2}. In the case of diffusion language models, all samples $\x_t$ seen at train time correspond directly to an embedding $\x_0$. However, this is not the case during inference, particularly for larger values of $t$, as the model reuses its own  (\Cref{fig:distance-to-closest-token}). This argument provides theoretical justification for the clamping trick proposed in \citealp{diffusionlm}, in which the prediction $\hat\x_0$ is clamped to the closest true embedding at every diffusion step.

% \subsection{Improving Inference Speed}

% \subsection{Architecture and Optimisation Details}

\section{Experiments}
\begin{figure}
    \centering
    \resizebox{0.9\linewidth}{!}{\input{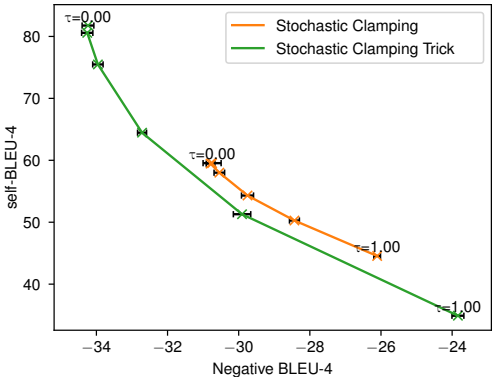}}
    \caption{Quality-diversity trade-off as the temperature $\tau$ is varied (lower is better for both metrics). Quality rapidly drops for $\tau>1$. Self-BLEU is measured over 5 seeds, and error bars denote standard deviation in quality. Note that the $\tau=0$ points correspond to the usual sampling procedures with and without the clamping trick.}
    \label{fig:qd-tradeoff-stochastic-clamping}
\end{figure}

\begin{table}[]
    \centering
    %\hspace*{-6cm}
    \resizebox{\linewidth}{!}{\centering\begin{tabular}{c|ccc|c}
    \hline
         & BLEU-4 & ROUGE-L & BERTScore & self-BLEU-4 \\
         \hline
         No clamping & 30.75 & 54.86 & 76.31 & \bf 59.54\\
         Clamping Trick & \bf 34.24 & \bf 57.97 & \bf 77.52 & 81.76\\
         % No clamping & $30.75 \pm 0.25$ & $54.86 \pm 0.03$ & $76.31 \pm 0.07$ & 59.54\\
         % Clamping Trick & $34.24 \pm 0.16$ & $57.97 \pm 0.21$ & $77.52 \pm 0.07$ & 81.76\\
         \hline
    \end{tabular}}
    \caption{Results on QQP with 5 inference steps with and without clamping. Self-BLEU is taken across 5 seeds.}
    \label{tab:clamp-noclamp-results}
\end{table}
%\end{multicols}

By aggregating techniques from recent methods and running several experiments, I was able to train a state-of-the-art language diffusion model for the paraphrasing task on the QQP dataset. This section proposes and evaluates methods for manipulating the model's diversity/quality trade-off.

\subsection{Controlling Diversity/Quality Trade-off}
\citealp{gans_fall_short} discuss at length the limitations of comparing language models based on their quality alone. Due to inherent quality-diversity trade-offs, comparisons must take into account model behaviour across a range of diversities rather than focusing on a single point.
% to compare language models fully it is necessary to approximate and compare their quality-diversity curves.

The quality-diversity trade-off of autoregressive models may naturally be controlled using sampling temperature, or more sophisticated methods such as stochastic beam search, while similar techniques exist for language GANs such as generator rejection sampling \cite{gans_fall_short}. In this vein, I explore three methods for controlling the diversity/quality trade-off at inference time:
\begin{enumerate}
    \item {\bf Stochastic Clamping.} Rather than clamp the final predicted embeddings $\hat\x_0$ to the closest tokens, a random selection is made from \emph{nearby} tokens.
    \item {\bf Classifier-free Guidance.} As observed elsewhere in the literature, a higher guidance strength should improve fidelity at the cost of reducing diversity \cite{classifier_free_guidance}.
    \item {\bf Combined Methods.} I explore methods for combining clamping and classifier-free guidance.
\end{enumerate}
% However, while some works do evaluate diversity along with quality \cite{diffuseq}, to the best of my knowledge no method for modulating the diversity-quality trade-off has yet been explored for diffusion language models.

\subsubsection{Stochastic Clamping} \label{sec:stochastic_clamping}
The final step of inference, given a predicted latent $\hat\x_0$, is usually
\[
    \hat\w_i = \argmax_{\w\in\wspace} || \Emb{\w} - \hat\x_{0, i} ||_2
\]
% (noting that $\hat\x_0$ has shape $S \times D$ for sequence length $S$ and model dimension $D$)
where $i$ gives the index in the sequence. This may be interpreted as a zero-temperature sample from a probability distribution with logits given by the distance
\[
    o(\w) = ||\Emb{\w} - \hat\x_{0, i}||_2
\]
for each index $i$. Inspired by the use of temperature in autoregressive models, we consider sampling from this distribution with non-zero temperature $\tau$ (with logits $o_\tau(\w)=o(\w)/\tau$), thus increasing diversity at the cost of quality.

\citealp{diffusionlm} propose the `clamping trick', in which the predictions $\hat\x_0$ are clamped to the nearest tokens at each diffusion step (rather than just once at the end), forcing the latents to remain close to actual token embeddings. This mitigates exposure bias, preventing $\hat\x_t$ from drifting away from the embeddings. While the clamping trick is beneficial for generation quality, it is highly detrimental to diversity (\Cref{tab:clamp-noclamp-results}) as the generation process commits earlier to a particular sequence. To remedy this, I propose the `stochastic clamping trick', in which stochastic clamping is used in place of hard clamping at every step of the diffusion process. The resulting quality-diversity trade-off is shown in \Cref{fig:qd-tradeoff-stochastic-clamping}, demonstrating that stochastic clamping enables the clamping trick to achieve superior quality with equal diversity to the baseline, and is therefore strictly beneficial for generation (using these particular metrics).
% demonstrating an improved trade-off for the stochastic clamping trick
% The clamping trick is therefore beneficial even when diversity is a priority, as 
% Note that the $\tau=0$ points on the two curves correspond to the usual sampling procedures with and without the clamping trick.

While stochastic clamping enables manipulation of the quality-diversity trade-off, increasing $\tau>0$ appears solely detrimental to quality, rendering it useful only for increasing diversity at the cost of quality.
% preventing this method from outperforming the baseline sampling method on quality alone.
% quality appears maximised for $\tau=0$, and so cannot be improved by increasing $\tau$.
% While stochastic clamping enables manipulation of the quality-diversity trade-off, it has a major limitation: quality appears maximised for $\tau=0$, and so cannot be improved by increasing $\tau$.
% so this method may not be used to improve generation quality.
% increasing $\tau$ above 0 appears strictly detrimental to quality, so this method may not be used to improve generation quality.

% This may be interpreted as a zero-temperature sample from the distribution
% Temperature-based sampling is not necessary for diffusion language models for diversity of samples

\begin{figure}
    \centering
    \resizebox{0.9\linewidth}{!}{\input{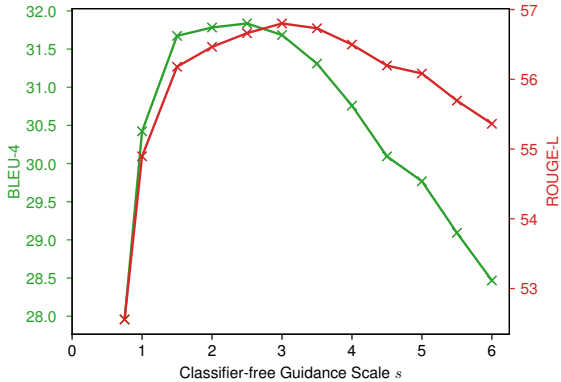}}
    \caption{Effect of classifier-free guidance scale on quality (BLEU-4 / ROUGE-L). Quality massively drops for $s<0.75$, omitted for clarity. Note that $s=1$ corresponds to the usual sampling procedure.}
    \label{fig:cfg-performance}
\end{figure}

\subsubsection{Classifier-free Guidance} \label{sec:cfg_experiments}
Classifier-free guidance has been shown to trade off diversity for quality in the domain of images up to a certain limit, beyond which the generation process becomes unstable \cite{classifier_free_guidance}. Evaluating across a range of guidance scales $s$, we also observe this pattern, with performance improving across several metrics for $s>1$ before dropping off (\Cref{fig:cfg-performance}).

% However, it is not clear that the usual classifier-free guidance method is the best fit for diffusion language models.
However, the discrete nature of text may pose a problem for the usual classifier-free guidance method. As the generation process progresses, the distance between the predicted embeddings $\hat\x_0$ and the closest true embedding $\Emb{\w}$ shrinks (\Cref{fig:distance-to-closest-token}). For small timesteps $t$, it becomes increasingly likely that both the conditional and unconditional prediction for a certain embedding fall very close to the same embedding $\Emb{\w}$. In such cases, classifier-free guidance only amplifies the error between the prediction and true embedding, potentially harming performance.

\begin{figure}
    \centering
    \resizebox{0.9\linewidth}{!}{\input{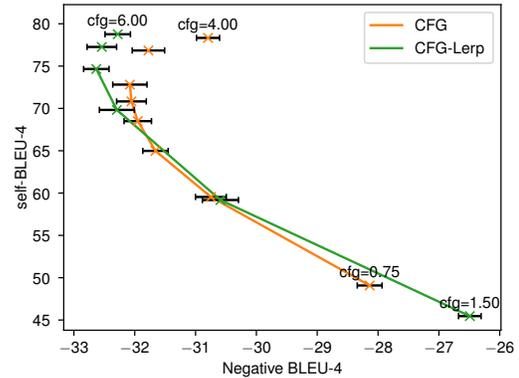}}
    \caption{Quality-diversity trade-off as the classifier-free guidance scale is varied. 5 seeds are used for self-BLEU and quality standard deviation error bars.}
    \label{fig:qd-tradeoff-cfg}
\end{figure}

We therefore explore two \emph{guidance schedules} for reducing the scale of the guidance as the diffusion process progresses. First, a simple linear schedule $s_t = \frac tTs$, and second the `standard deviation' schedule $s_t = \sqrt{1-\overline \alpha_t}$, in which $s_t$ equals the standard deviation of the forward process $q(\x_t \mid \x_0)$ taken from the closed form in \cref{eq:closed_form}. \Cref{fig:qd-tradeoff-cfg} indicates the quality-diversity trade-off as $s$ is varied for both the usual CFG and the interpolated schedule (`CFG-Lerp') - we find that the linear schedule achieves a strictly superior trade-off compared to the standard deviation schedule (on both BLEU-4 and ROUGE-L), and so omit the latter from figures for clarity. \Cref{fig:qd-tradeoff-cfg} confirms the hypothesis that a classifier-free guidance scale greater than 1 improves quality at the cost of reducing diversity in diffusion language models, as observed previously in image models. Furthermore, the interpolated schedule enables higher guidance scales, and ultimately achieves a higher peak quality, but does not strictly dominate the constant schedule.

\Cref{fig:distance-to-closest-token} provides evidence for the claim that a constant guidance scale exacerbates prediction errors at lower timesteps; the distance to the nearest token remains relatively high as $t$ approaches $0$ for the $s=2.5$ line (green), whereas the interpolated schedule (red) achieves a low final distance matching that of the baseline.

While the guidance schedule $s_{1..T}$ should depend directly on the noise schedule $\beta_{1..T}$, we establish a linear schedule as a baseline and leave further investigation to future work.

% I therefore explore using a \emph{schedule} for the guidance, reducing the scale as the diffusion process progresses. Two options
% \[
%     s_t = \frac{t}{T}s
% \]
% as well as a
% applied to this prediction, for larger scales $s$
% when using a scale above 1
% observe quality improvements across several metrics using classifier-free guidance

% TODO: CFG / performance graph, put BLEU and ROUGE on there?
%  - maybe put CFG-LERP on there too
% TODO: Explain + justify CFG-LERP
% TODO: DIV-QAL graph for CFG + CFG-LERP

% \begin{figure}
%     \centering
%     \resizebox{0.9\linewidth}{!}{\input{graphs/cfg-lerp-performance.pgf}}
%     \caption{Effect of classifier-free guidance scale on quality (BLEU-4 / ROUGE-L) when using a linearly interpolated scale.}
%     \label{fig:cfg-lerp-performance}
% \end{figure}

\subsubsection{Combined Methods}
The previous sections establish (stochastic) clamping (\cref{sec:stochastic_clamping}) and classifier-free guidance (\cref{sec:cfg_experiments}) as inference-time methods to improve quality and control the quality-diversity trade-off. This section investigates the possibility of combining these independent methods.

There are several options to explore for the combination of classifier-free guidance and the clamping trick. The simplest options are `cfg-before-clamp' (\ref{eq:cfg-before-clamp}) and `clamp-before-cfg' (\ref{eq:clamp-before-cfg}):
\begin{align}
    \predtilde &= \clamp(\cfg(\predu, \predc)) \label{eq:cfg-before-clamp} \\
    \predtilde &= \cfg(\clamp(\predu), \clamp(\predc)) \label{eq:clamp-before-cfg}
\end{align}
using the notation $f_\theta(\x_t, t, \cond) = \predc$ for the conditional prediction, $f_\theta(\x_t, t, \emptyset) = \predu$ for the unconditional prediction, $\clamp$ for the (possibly stochastic) clamping operation and $\cfg$ for the classifier-free guidance operation (possibly with a non-constant schedule $s_t$).

\begin{figure}
    \centering
    % \resizebox{0.9\linewidth}{!}{ \input{graphs/model-output-dist-to-tok.pgf}}
    \resizebox{0.9\linewidth}{!}{ \input{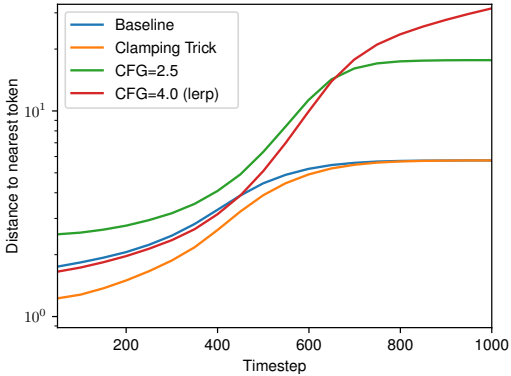}}
    \caption{Distance of predicted embeddings $\hat\x_0$ (following CFG) to closest true embedding during 20-step inference, averaged over evaluation on the entire test set. Note the logarithmic scale on the y-axis due to significant differences in scale. `Baseline' refers to the usual inference procedure without clamping or CFG. The strength values of $s=2.5$ and $s=4.0$ were chosen to maximise BLEU score.}
    \label{fig:distance-to-closest-token}
\end{figure}

\begin{figure}
    \centering
    % \resizebox{0.9\linewidth}{!}{ \input{graphs/model-output-dist-to-tok.pgf}}
    \resizebox{0.9\linewidth}{!}{ \input{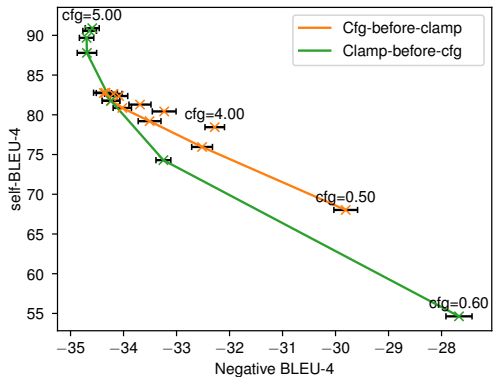}}
    \caption{Quality-diversity trade-off for clamp-before-cfg and cfg-before-clamp, measured over 5 runs.}
    \label{fig:qd-tradeoff-CBC}
\end{figure}

One might argue that model outputs $\predu$, $\predc$ should always equal $\Emb{\w}$ for some $\w\in\wspace$, and that any difference is erroneous. This provides justification for the `clamp-before-cfg' method; the model's output should be clamped to the nearest token to minimise the errors before they are exacerbated by classifier-free guidance. \Cref{fig:distance-to-closest-token} casts some doubt on this claim; even when using clamping, model outputs (for large $t$) may have moderate distance to the nearest token $\w$ (mean distance is over 5 for $t=1000$). However, the dramatic distance increase when using CFG provides a strong argument against `cfg-before-clamp' - the model outputs following CFG are often very far from the nearest embedding, causing clamping to significantly change the prediction in an unpredictable manner. This explains the result in \Cref{fig:qd-tradeoff-CBC}, in which `clamp-before-cfg' achieves a strictly superior trade-off to `cfg-before-clamp'.

% GRAPH TODO
%  - Length / performance graph
%  - Fidelity / quality tradeoff graph as LENGTH changes
% - Graph of final performances compared to alternative methods
%  - DiffuSeq
%  - D3PM ?
%  - Difformer
%  - Base transformer
% - Talk about / measure inference speed

\subsubsection{Efficiency}
Temperature based sampling, in both autoregressive models and stochastic sampling, is essentially free of cost, adding little overhead compared to the hard sampling alternative. Beam search and Minimum Risk Bayes (MBR) decoding \cite{minimum-bayes-risk-decoding}, on the other hand, are highly expensive, requiring a linearly increasing amount of computation with respect to the beam width / set size. Classifier-free guidance is cheap compared to MBR, with a fixed cost of doubling the number of model evaluations making it more efficient than MBR for $|S| \geq 2$. The clamping trick requires additional clamping operations, but as these are highly parallelisable, and the number of inference timesteps is generally small, the overhead is small compared to classifier-free guidance.

% Classifier-free guidance and clamping are very efficient compared to beam search and MBR, but do require some overhead; the clamping trick requires an expensive clamping operation to be carried out $K$ times, where $K$ is the number of inference diffusion steps, while classifier-free guidance doubles the number of model evaluations (though this overhead is constant with respect to the scale $s$). Classifier-free guidance is strictly more efficient than running inference twice (as the clamping step need only be carried out once), and is therefore more efficient than MBR for $|S|\geq 2$. I find the overhead from the clamping trick to be relatively small due to the small value of inference steps $K=20$ and as clamping is a highly parallelisable task. 

The clamp-before-cfg method is the most expensive presented here, requiring two clamping operations per diffusion step, as well as doubling the number of model evaluations. However, this is still far more efficient than MBR for large set sizes, such as $|S|=10$ in \citealp{diffuseq}. Overall, the proposed methods are not without overhead, but this overhead is constant with respect to strength $s$ and small compared to the commonly used MBR.

% However, as before, it is strictly more efficient than running inference using the clamping trick twice, and is therefore faster than MBR (with the clamping trick) for $|S|\geq 2$. Overall, the proposed methods are fairly expensive, approximately doubling inference time, but faster than MBR with a large set size (e.g. $|S|=10$ is used in \citealp{diffuseq}).

% C TRICK: N clamp + N model + 1
% CBC:    2N clamp + 2N model + 1

% Both clamping and classifier-free guidance incur an overhead, but importantly this overhead is \emph{constant} with respect to the temperature

\subsection{Controllable Length}
\Cref{tab:examples-best-quality} gives several example outputs at different lengths, showing the model's ability to produce reasonable texts for a range of lengths. It is interesting to note that even for inference methods with very low diversity (such as those in \cref{tab:examples-best-quality}), varying the length can produce quite different generations. This may be significant when applying MBR across lengths, as is done in \cite{difformer}, as this provides a more significant performance increase when diversity is high. While the model may output samples of any length, performance significantly drops for longer outputs and token repetition becomes extremely common.

% Naturally, performance significantly deteriorates when length is increased above a reasonable level. For the condition \textit{How do I learn a computer language like java?}, a 50-token response with the baseline 20-step model is
% \begin{center}
%     \textit{what is good way to must learn using java computer and i i like i java? what, with i by learn programming language and'can like learning, language being ', is java language future? java with java java java what? is? java}
% \end{center}
% highlighting a common issue at larger lengths - token repetition. 

% TABLE 2: demonstrate sample diversity for various CFGs
% Input     Model1      Model2      Model3
% adasd     ...         ...         ...     
%           ...         ...         ...
%           ...         ...         ...
% adasd     ...         ...         ...     
%           ...         ...         ...
%           ...         ...         ...
\begin{table*} % cfg = 0.5, 1.0, 2.0, 3.0
    \centering
    \hspace*{-0.4cm}
    \resizebox{1.05\linewidth}{!}{
    \begin{tabular}{m{5cm}|c|c|c}
         \bf Input & \multicolumn{3}{c}{\bf Generated text}  \\
%         \hline
         ~ & $s=0.5$ & $s=1.0$ & $s=2.5$ \\
         \hline
         How do I make money with YouTube?  & \makecell{how can i have money through youtube?\\how can i make money through youtube?\\how do i make money fast from?\\how can we make money from youtube?\\how can i make money the youtube?} & \makecell{how can i make money on youtube?\\how can i make money on youtube?\\how can i make money on youtube?\\how can you make money through youtube?\\how can you make money from youtube?} & \makecell{how can i make money through youtube?\\how can people make money on youtube?\\how can people earn money on youtube?\\how can i make money from youtube?\\how can i make money from youtube?} \\ \hline
          How do I become a good computer science engineer? & \makecell{how can i good for an engineer?\\how can i be good on online?\\how do i good for science engineer?\\how can i a be on us?\\how do i be the computer science?} & \makecell{how can i be good science engineer?\\how do i a good science engineer?\\how do i good computer science engineer?\\how can i be good science engineer?\\how do i be computer science engineer?} & \makecell{how can i be good computer engineer?\\how can i be good science engineer?\\how do i good computer science engineer?\\how do i good computer science engineer?\\how do i be good science engineer?} \\ \hline
          I was suddenly logged off Gmail. I can't remember my Gmail password and just realized the recovery email is no longer alive. What can I do? & \makecell{how do i lose my writing password?\\how do i not i to email?\\where can i recover my gmail?\\how can i lose gma from account?\\how do i l very from password?} & \makecell{how can i reset my gma password?\\how do i recover gmail password?\\how do i reset youril password?\\how can i reset gmail password?\\how do i reset my gma password?} & \makecell{how do i reset gmail password?\\how do i reset gmail password?\\how do i reset gmail password?\\how do i my gmail password?\\how do i reset myil password?}
    \end{tabular}}
    \caption{Example outputs using classifier-free guidances scales $s\in\{0.5,1.0,2.5\}$. All outputs have a \emph{fixed length of 8 tokens} and use 20 DPM++ steps. Notice how as $s$ increases, relevance to the condition (and quality) increases while diversity decreases. $s=2.5$ is the value for which BLEU score is maximised. The bottom example exemplifies an innate support for text summarisation. Finally, the middle row is a good example of the model's inability to choose the length of its output being detrimental to performance; it has insufficient tokens to complete the sentence.}
    \label{tab:examples-cfg-vary}
\end{table*}

% TABLE 1: demonstrate length control
% Input     Model1      Model2      Model3
% adasd     ...         ...         ...     
%           ...         ...         ...
%           ...         ...         ...
% adasd     ...         ...         ...     
%           ...         ...         ...
%           ...         ...         ...
\begin{table*} % cfg = 0.5, 1.0, 2.0, 3.0
    \centering
    \hspace*{-0.4cm}
    \resizebox{1.05\linewidth}{!}{
    \begin{tabular}{m{5cm}|c|c}
         \bf Input & \multicolumn{2}{c}{\bf Generated text}  \\
%         \hline
         ~ & Clamping Trick & Clamp-before-cfg, $s=3$ \\ \hline
         How do I make friends? & \makecell{how make friends?\\how do i make friends?\\how do i make friends with.?\\what is the best way to make a friend?\\i do i friends and how should i do about it?} & \makecell{how we friends?\\how do we make friends?\\how do i make friends with friend?\\what is the best way to make a friend?\\i have no friends and how should i do about it?}\\ \hline
         What movies would you recommend everyone to watch?  & \makecell{what to watch?\\what movies can to watch?\\what movies would you recommend to watch?\\what are would you movies to recommend to watch?\\what is the one movies you watch to recommend to watch?} & \makecell{what you watch?\\what movies you to watch?\\what movies would you recommend to watch?\\what are the you movies to recommend to watch?\\what is the one movie that you to recommend to watch?}
    \end{tabular}}
    \caption{Examples outputs with fixed lengths of [4, 6, 8, 10, 12] tokens respectively, using the two inference methods with the highest measured quality. 20 generation steps are used for each sample.}
    \label{tab:examples-best-quality}
\end{table*}

\subsection{Comparison to State-of-the-art}
% TODO could put these side-by-side whole page
\begin{figure}
    \centering
    \resizebox{0.9\linewidth}{!}{\input{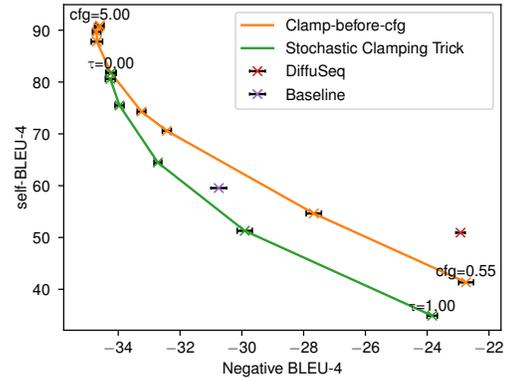}}
    \caption{Quality-diversity trade-off using BLEU-4 over 5 runs for the best performing methods.}
    \label{fig:qd-big-bleu}
\end{figure}

\begin{table*}
    \centering
    \hspace*{-0.6cm}
    \begin{tabular}{l|c|cccc|c}
        \bf Model & \bf Steps & \bf BLEU-4 & \bf BLEU-4$_\text{nltk}$ & \bf ROUGE-L & \bf BERTScore & \bf self-BLEU \\
        \hline
        Fine-tuned GPT-2 & - & - & 20.59 & - & \bf 83.63 & -\\
        \hline
        Difformer\footnotemark \cite{difformer} & 20 & - & 30.43 & 61.25 & 82.22 & -  \\
        DiffuSeq \cite{diffuseq} & 2000 & 22.92 & 20.37 & 50.82 & 79.20 & 50.93 \\
        SeqDiffuSeq \cite{SeqDiffuSeq} & 2000 & - & 23.28 & - & 82.91 & - \\
        \hline
        \name & 20 & 28.86 & 27.04 & 53.53 & 75.50 & 55.42 \\
        \name~($s=0.75$) & 20 & 25.79 & 24.07 & 50.43 & 73.44 & \bf 44.24 \\
        \name~(CT) & 20 & 34.17 & 32.49 & 57.82 & 77.68 & 78.39  \\
        \name~(CT, $s=3$) & 20 & \bf 34.81 & \bf 33.17 & \bf 58.33 & 77.74 & 88.64
    \end{tabular}
    \caption{Evaluation results on QQP, with best shown in bold. `CT' denotes the clamping trick, and the final row uses clamp-before-cfg to combine cfg and clamping. Fine-tuned GPT-2 and SeqDiffuSeq results are quoted from SeqDiffuSeq, Difformer results quoted from their GitHub, and DiffuSeq results are measured from their provided outputs. Full evaluation details are in \cref{sec:eval-metrics}.}
    \label{tab:big-comparison}
\end{table*}
\footnotetext{MBR of unspecified size over multiple lengths. Best results not bolded as it is unfair to compare to the other methods with have MBR size 1.}

\begin{figure}
    \centering
    \resizebox{0.9\linewidth}{!}{\input{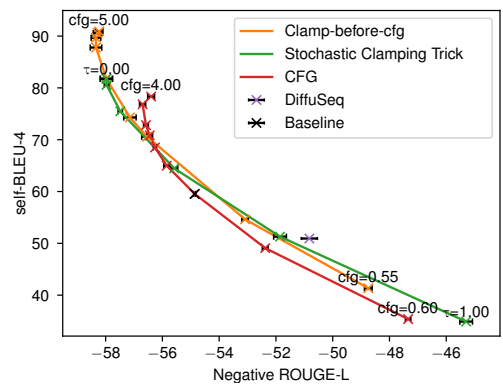}}
    \caption{Quality-diversity trade-off using ROUGE-L over 5 runs for the best performing methods.}
    \label{fig:qd-big-rouge}
\end{figure}

\Cref{fig:qd-big-bleu} and \Cref{fig:qd-big-rouge} demonstrate the final BLEU-4 and ROUGE-L quality-diversity trade-offs of my model. Both figures display only the inference methods which are not strictly dominated by any other method. While many relevant past works exist, few evaluate diversity using self-BLEU, and there are several different implementations of BLEU/ROUGE-L used, so we compare only to DiffuSeq \cite{diffuseq}\footnote{DiffuSeq released model outputs as text so we evaluate on these directly rather than using the values in the paper due to implementation differences. We do not use MBR (as in the original) for fairness of comparison.} in these figures. While the baseline performance does not dominate DiffuSeq, due to reduced diversity, it is clear that the overall trade-off curve is superior.

\Cref{tab:big-comparison} gives a comparison of several inference procedures to existing methods, and demonstrates superior performance (without MBR) on the BLEU, ROUGE and self-BLEU metrics. However, it is important to note that my measured BLEU and ROUGE scores are \emph{artificially inflated} compared to other methods due to the way we handle length - further discussion in Limitations. This has a lesser effect on the semantic BERTScore, leading to more accurate scores. While maximising performance was not the goal of this project, generation quality is currently limited due to use of the BERT tokenizer, rather than one specialized for QQP, as well as the model's inability to choose the length of its output - the latter issue is mitigated using MBR across multiple lengths in Difformer \cite{difformer}, which is the only other model in \Cref{tab:big-comparison} with controllable length. 

\section{Conclusions} \label{sec:conclusions}
This project proposes several novel inference methods for manipulating the quality-diversity trade-off in diffusion language models: classifier-free guidance, stochastic sampling and the clamp-before-cfg method. We train a baseline model and apply these methods to validate their efficacy, achieving performance competitive with the current state-of-the-art despite training for just three hours. We demonstrate that the proposed methods not only enable high diversity samples, but also, applied in the opposite direction, significantly improve performance over the baseline (+5.9 BLEU-4, +4.8 ROUGE-L and +2.2 BERTScore points) despite being relatively cheap modifications to inference. Overall, this project shows that diffusion language models are capable of generating text at a wide range of diversities and qualities as with autoregressive language models, and propose several methods for achieving this efficiently.

\section*{Limitations} \label{sec:limitations}
% TODO: Put a bunch of ROUGE-L graphs in an appendix and link here
Computation limitations reduced my ability to calculate BERTScore, so most evaluation is done using BLEU, despite its drawbacks \cite{bleu-review-reiter, need-new-eval-metrics-nlg}. Due to time constraints we have been unable to produce a performant length model, and therefore use the \emph{length of the ground truth outputs} as the lengths of my predictions during evaluation - this inflates performance. However, this does not subtract from the key takeaways of this project of creating a method for manipulating the quality-diversity trade-off for diffusion language models.

% The BLEU metric is widely used to enable comparisons with past work, despite its many issues \cite{bleu-review-reiter, need-new-eval-metrics-nlg}. 

It would've been interesting to compare the quality-diversity trade-off to that of an autoregressive model, but as QQP is rarely used for generative tasks we were not able to find a suitable pre-trained transformer. This analysis being limited to QQP is a further limitation - with more time and compute we would like to apply this to other datasets/tasks, such as text summarisation in which the controllable length has particular value.

% \section*{Ethics Statement} % what

% \section*{Acknowledgements}

\bibliographystyle{ieeenat_fullname}
\bibliography{main}

\begin{thebibliography}{29}
\providecommand{\natexlab}[1]{#1}
\providecommand{\url}[1]{\texttt{#1}}
\expandafter\ifx\csname urlstyle\endcsname\relax
  \providecommand{\doi}[1]{doi: #1}\else
  \providecommand{\doi}{doi: \begingroup \urlstyle{rm}\Url}\fi

\bibitem[Anonymous(2023{\natexlab{a}})]{diffusion_exposure_bias_2}
Anonymous.
\newblock Alleviating exposure bias in diffusion models through sampling with shifted time steps.
\newblock In \emph{Submitted to The Twelfth International Conference on Learning Representations}, 2023{\natexlab{a}}.
\newblock under review.

\bibitem[Anonymous(2023{\natexlab{b}})]{reparameterized_discrete_diffusion_model}
Anonymous.
\newblock A reparameterized discrete diffusion model for text generation.
\newblock In \emph{Submitted to The Twelfth International Conference on Learning Representations}, 2023{\natexlab{b}}.
\newblock under review.

\bibitem[Austin et~al.(2021)Austin, Johnson, Ho, Tarlow, and van~den Berg]{D3PM}
Jacob Austin, Daniel~D. Johnson, Jonathan Ho, Daniel Tarlow, and Rianne van~den Berg.
\newblock Structured denoising diffusion models in discrete state-spaces.
\newblock In \emph{Advances in Neural Information Processing Systems}, pages 17981--17993. Curran Associates, Inc., 2021.

\bibitem[Caccia et~al.(2020)Caccia, Caccia, Fedus, Larochelle, Pineau, and Charlin]{gans_fall_short}
Massimo Caccia, Lucas Caccia, William Fedus, Hugo Larochelle, Joelle Pineau, and Laurent Charlin.
\newblock Language gans falling short, 2020.

\bibitem[Chen et~al.(2021)Chen, Zhang, Zen, Weiss, Norouzi, and Chan]{WaveGrad_audio_diffusion}
Nanxin Chen, Yu Zhang, Heiga Zen, Ron~J Weiss, Mohammad Norouzi, and William Chan.
\newblock Wavegrad: Estimating gradients for waveform generation.
\newblock In \emph{International Conference on Learning Representations}, 2021.

\bibitem[Dhariwal and Nichol(2021)]{diffusion_beats_gans}
Prafulla Dhariwal and Alexander Nichol.
\newblock Diffusion models beat gans on image synthesis.
\newblock \emph{Advances in neural information processing systems}, 34:\penalty0 8780--8794, 2021.

\bibitem[Eikema and Aziz(2020)]{map-all-you-need}
Bryan Eikema and Wilker Aziz.
\newblock Is {MAP} decoding all you need? the inadequacy of the mode in neural machine translation.
\newblock In \emph{Proceedings of the 28th International Conference on Computational Linguistics}, pages 4506--4520, Barcelona, Spain (Online), 2020. International Committee on Computational Linguistics.

\bibitem[Gao et~al.(2023)Gao, Guo, Tan, Zhu, Zhang, Bian, and Xu]{difformer}
Zhujin Gao, Junliang Guo, Xu Tan, Yongxin Zhu, Fang Zhang, Jiang Bian, and Linli Xu.
\newblock Difformer: Empowering diffusion models on the embedding space for text generation, 2023.

\bibitem[Gong et~al.(2023)Gong, Li, Feng, Wu, and Kong]{diffuseq}
Shansan Gong, Mukai Li, Jiangtao Feng, Zhiyong Wu, and Lingpeng Kong.
\newblock Diffuseq: Sequence to sequence text generation with diffusion models, 2023.

\bibitem[Ho and Salimans(2022)]{classifier_free_guidance}
Jonathan Ho and Tim Salimans.
\newblock Classifier-free diffusion guidance, 2022.

\bibitem[Ho et~al.(2020)Ho, Jain, and Abbeel]{DDPM}
Jonathan Ho, Ajay Jain, and Pieter Abbeel.
\newblock Denoising diffusion probabilistic models.
\newblock In \emph{Advances in Neural Information Processing Systems}, pages 6840--6851. Curran Associates, Inc., 2020.

\bibitem[Kong et~al.(2021)Kong, Ping, Huang, Zhao, and Catanzaro]{DiffWave_audio_diffusion}
Zhifeng Kong, Wei Ping, Jiaji Huang, Kexin Zhao, and Bryan Catanzaro.
\newblock Diffwave: A versatile diffusion model for audio synthesis.
\newblock In \emph{International Conference on Learning Representations}, 2021.

\bibitem[Kumar and Byrne(2004)]{minimum-bayes-risk-decoding}
Shankar Kumar and William Byrne.
\newblock Minimum {B}ayes-risk decoding for statistical machine translation.
\newblock In \emph{Proceedings of the Human Language Technology Conference of the North {A}merican Chapter of the Association for Computational Linguistics: {HLT}-{NAACL} 2004}, pages 169--176, Boston, Massachusetts, USA, 2004. Association for Computational Linguistics.

\bibitem[Li et~al.(2022)Li, Thickstun, Gulrajani, Liang, and Hashimoto]{diffusionlm}
Xiang~Lisa Li, John Thickstun, Ishaan Gulrajani, Percy Liang, and Tatsunori~B. Hashimoto.
\newblock Diffusion-lm improves controllable text generation, 2022.

\bibitem[Liu et~al.(2022)Liu, Ren, Lin, and Zhao]{PNDM}
Luping Liu, Yi Ren, Zhijie Lin, and Zhou Zhao.
\newblock Pseudo numerical methods for diffusion models on manifolds.
\newblock In \emph{International Conference on Learning Representations}, 2022.

\bibitem[Lu et~al.(2023)Lu, Zhou, Bao, Chen, Li, and Zhu]{dpm_solver_plusplus}
Cheng Lu, Yuhao Zhou, Fan Bao, Jianfei Chen, Chongxuan Li, and Jun Zhu.
\newblock Dpm-solver++: Fast solver for guided sampling of diffusion probabilistic models, 2023.

\bibitem[Nichol et~al.(2021)Nichol, Dhariwal, Ramesh, Shyam, Mishkin, McGrew, Sutskever, and Chen]{GLIDE}
Alex Nichol, Prafulla Dhariwal, Aditya Ramesh, Pranav Shyam, Pamela Mishkin, Bob McGrew, Ilya Sutskever, and Mark Chen.
\newblock Glide: Towards photorealistic image generation and editing with text-guided diffusion models.
\newblock In \emph{International Conference on Machine Learning}, 2021.

\bibitem[Ning et~al.(2023)Ning, Li, Su, Salah, and Ertugrul]{diffusion_exposure_bias}
Mang Ning, Mingxiao Li, Jianlin Su, Albert~Ali Salah, and Itir~Onal Ertugrul.
\newblock Elucidating the exposure bias in diffusion models.
\newblock In \emph{Submitted to The Twelfth International Conference on Learning Representations}, 2023.

\bibitem[Novikova et~al.(2017)Novikova, Du{\v{s}}ek, Cercas~Curry, and Rieser]{need-new-eval-metrics-nlg}
Jekaterina Novikova, Ond{\v{r}}ej Du{\v{s}}ek, Amanda Cercas~Curry, and Verena Rieser.
\newblock Why we need new evaluation metrics for {NLG}.
\newblock In \emph{Proceedings of the 2017 Conference on Empirical Methods in Natural Language Processing}, pages 2241--2252, Copenhagen, Denmark, 2017. Association for Computational Linguistics.

\bibitem[Reiter(2018)]{bleu-review-reiter}
Ehud Reiter.
\newblock A structured review of the validity of {BLEU}.
\newblock \emph{Computational Linguistics}, 44\penalty0 (3):\penalty0 393--401, 2018.

\bibitem[Rombach et~al.(2022)Rombach, Blattmann, Lorenz, Esser, and Ommer]{latent_diffusion_models}
R. Rombach, A. Blattmann, D. Lorenz, P. Esser, and B. Ommer.
\newblock High-resolution image synthesis with latent diffusion models.
\newblock In \emph{2022 IEEE/CVF Conference on Computer Vision and Pattern Recognition (CVPR)}, pages 10674--10685, Los Alamitos, CA, USA, 2022. IEEE Computer Society.

\bibitem[Song et~al.(2021)Song, Meng, and Ermon]{DDIM}
Jiaming Song, Chenlin Meng, and Stefano Ermon.
\newblock Denoising diffusion implicit models.
\newblock In \emph{International Conference on Learning Representations}, 2021.

\bibitem[Song et~al.(2023)Song, Dhariwal, Chen, and Sutskever]{consistency}
Yang Song, Prafulla Dhariwal, Mark Chen, and Ilya Sutskever.
\newblock Consistency models, 2023.

\bibitem[Stahlberg and Byrne(2019)]{cat-got-your-tongue}
Felix Stahlberg and Bill Byrne.
\newblock On {NMT} search errors and model errors: Cat got your tongue?
\newblock In \emph{Proceedings of the 2019 Conference on Empirical Methods in Natural Language Processing and the 9th International Joint Conference on Natural Language Processing (EMNLP-IJCNLP)}, pages 3356--3362, Hong Kong, China, 2019. Association for Computational Linguistics.

\bibitem[Vaswani et~al.(2017)Vaswani, Shazeer, Parmar, Uszkoreit, Jones, Gomez, Kaiser, and Polosukhin]{transformer}
Ashish Vaswani, Noam Shazeer, Niki Parmar, Jakob Uszkoreit, Llion Jones, Aidan~N Gomez, \L~ukasz Kaiser, and Illia Polosukhin.
\newblock Attention is all you need.
\newblock In \emph{Advances in Neural Information Processing Systems}. Curran Associates, Inc., 2017.

\bibitem[von Platen et~al.(2022)von Platen, Patil, Lozhkov, Cuenca, Lambert, Rasul, Davaadorj, and Wolf]{diffusers}
Patrick von Platen, Suraj Patil, Anton Lozhkov, Pedro Cuenca, Nathan Lambert, Kashif Rasul, Mishig Davaadorj, and Thomas Wolf.
\newblock Diffusers: State-of-the-art diffusion models.
\newblock \url{https://github.com/huggingface/diffusers}, 2022.

\bibitem[Yuan et~al.(2022)Yuan, Yuan, Tan, Huang, and Huang]{SeqDiffuSeq}
Hongyi Yuan, Zheng Yuan, Chuanqi Tan, Fei Huang, and Songfang Huang.
\newblock Seqdiffuseq: Text diffusion with encoder-decoder transformers.
\newblock \emph{ArXiv}, abs/2212.10325, 2022.

\bibitem[Zhang et~al.(2023)Zhang, Liu, and Zhang]{diffusum}
Haopeng Zhang, Xiao Liu, and Jiawei Zhang.
\newblock Diffusum: Generation enhanced extractive summarization with diffusion, 2023.

\bibitem[Zhu et~al.(2018)Zhu, Lu, Zheng, Guo, Zhang, Wang, and Yu]{Texygen}
Yaoming Zhu, Sidi Lu, Lei Zheng, Jiaxian Guo, Weinan Zhang, Jun Wang, and Yong Yu.
\newblock Texygen: A benchmarking platform for text generation models.
\newblock In \emph{The 41st International ACM SIGIR Conference on Research \& Development in Information Retrieval}, page 1097–1100, New York, NY, USA, 2018. Association for Computing Machinery.

\end{thebibliography}

\appendix

\onecolumn
\crefalias{section}{appendix}

\section{Evaluation Metrics} \label{sec:eval-metrics}
We use the following implementations:
\begin{itemize}
    \item BLEU-4: \verb|https://github.com/mjpost/sacrebleu|\footnote{hash \texttt{"nrefs:1|case:lc|eff:no|tok:13a|smooth:exp|version:2.4.0}} - we additionally calculate \verb|nltk| scores in one place, and this is clearly marked.
    \item ROUGE-L: \verb|https://github.com/pltrdy/rouge|
    \item BERTScore: \verb|https://github.com/Tiiiger/bert_score|\footnote{hash \texttt{microsoft/deberta-xlarge-mnli\_L40\_no-idf\_version=0.3.12(hug\_trans=4.34.1)}}
    \item self-BLEU: we could not find a standard implementation of self-BLEU, and in fact observed very significant differences between the versions used in different papers. We therefore implement our own version using \verb|sacrebleu| (detailed in the following section).
\end{itemize}

\subsection{self-BLEU}
Given a set of hypotheses $S$, self-BLEU is defined as the mean corpus-BLEU with hypothesis-reference pairs $(s, S\setminus\{s\})$ for each $s\in S$. Formally,
\[
    \text{self-BLEU}(S) = \frac1{|S|}\sum_{s \in S} \text{corpus\_bleu}(s, S \setminus \{s\})
\]
using \verb|sacrebleu|'s implementation of \verb|corpus_bleu|. For each condition in the test set, $|S|=5$ hypotheses are generated to form $S$, and $\text{self-BLEU}(S)$ is calculated - we report the mean value of $\text{self-BLEU}(S)$ over all conditions. This follows the original proposal in \cite{Texygen}, although their reference implementation (possibly erroneously) includes $s$ in the reference set for itself, leading to very high self-BLEU values.

\section{Hyperparameters} \label{sec:hyperparameters}
We use a transformer encoder-decoder model with 12 encoder/decoder layers, 12 attention heads, and dimension 784. We use the BERT tokenizer, and find that jointly learning embeddings during training outperforms using the pre-trained BERT embeddings as previously observed in \citealp{diffuseq, difformer}. The embeddings are 128-dimensional, and inputs/outputs are projected to/from the transformer dimension via a single linear layer. We use the AdamW optimiser, a cosine learning rate schedule from 1e-4 to 0 and find it necessary to use 2000 warmup steps to prevent a form of modal collapse in which the model always outputs a single token. Training on QQP for 100 epochs using mixed precision and \verb|torch.compile| takes just 3 hours on a single A100 80GB GPU.

$T=1000$ training timesteps are used, with DPM++ used for inference. The $\beta$ schedule is linear, and interpolates from 0.02 ($t=T$) to 0.0001.

\section{Unconditional Outputs}
Unconditional outputs may be achieved by setting the CFG scale $s=0$. As $p=0.1$, unconditional outputs occupy a small proportion of the model's capacity and are therefore generally of lower quality. Below are a large number of non-cherry picked unconditional outputs of fixed length 8 for the baseline model with 20 steps:
\begin{center}
\begin{BVerbatim}
what would i for website in account?
what is the best way of men?
how does happen help one get people?
what are the best books for bangalore?
how can i start to my easily?
who are some examples of " life?
how do i become to best website?
what are best ways to view account?
what are some ways of your life?
how do i start with quora?
how can i earn moneya english?
who is happen to best love for?
what do you safe of your life?
how can you of president your universe?
how can i like paper someone travel?
how do i become money changed not?
how can i get a book phone?
is there any bad like to data?
how do you love your travel out?
which is the most important rest engineer?
how do you become a worst??
how do i become a test online?
how can you earn a hard startup?
who should a learning it? why?
what is the best companies in india?
how do i improve my english weight?
how do you get on quora?
is it possible to know for 2000?
what are some of days weight movies?
what are the best of at life?
what is the best way to learn?
what is the way to learn you?
why should i create money at online?
how can i improve my quora?
is there any un good ar?
what is some col has to life?
how i can books people for english?
how do i feel get in phone?
how is oldest websites of service 2016?
what is job of iphone and why?
what i should use for chinese english?
do did water good right to one?
who is a worst my out world?
what are some top 10 in india?
what is the best of international language?
how can we earn money your famous?
what is the best books for 2017?
what are the point of face programming?
what are it as your feel fast?
what are some ways to score online?
what should i do to become better?
what is the important math of 50?
what can be your purpose at english?
\end{BVerbatim}
\end{center}

\end{document}